\documentclass{article}
\usepackage{iclr_template/iclr2027_conference,times}

\usepackage{amsmath,amsfonts,bm}

\def\eqref#1{equation~\ref{#1}}
\def\1{\bm{1}}

\DeclareMathAlphabet{\mathsfit}{\encodingdefault}{\sfdefault}{m}{sl}
\SetMathAlphabet{\mathsfit}{bold}{\encodingdefault}{\sfdefault}{bx}{n}

\usepackage{graphicx}
\usepackage{hyperref}
\usepackage{url}
\usepackage{amsmath}
\usepackage{amssymb}
\usepackage{mathtools}
\usepackage{amsthm}
\usepackage{arydshln}
\usepackage{algorithm}
\usepackage{algorithmic}
\usepackage{xcolor}
\usepackage{tikz}
\usetikzlibrary{shapes.geometric, arrows, positioning, fit, backgrounds, matrix, arrows.meta, shadows, decorations.pathreplacing}
\usepackage{pgfplots}
\pgfplotsset{compat=1.18}
\usepackage{soul}
\usepackage{tcolorbox}
\usepackage{multirow}
\usepackage[capitalize,noabbrev]{cleveref}
\usepackage{subcaption}
\usepackage{wrapfig}
\usepackage[textsize=tiny]{todonotes}
\usepackage{newfloat}
\usepackage{listings}
\usepackage{booktabs}

\DeclareCaptionStyle{ruled}{labelfont=normalfont,labelsep=colon,strut=off}
\floatstyle{ruled}
\newfloat{listing}{tb}{lst}{}
\floatname{listing}{Listing}

\iclrfinalcopy

\title{RRC: Unlocking Generative Reward Models in LLM Reinforcement Learning via Ranking-Based Reward Construction}
\author{
Chenglong Wang$^{1,2}$,
Ziming Zhu$^1$,
Yifu Huo$^1$,
Bei Li$^1$,
Qiaozhi He$^1$,
Yan Ding$^1$, \\
\textbf{
Xiaoyang Hao$^1$, 
Yuxin Gao$^1$,
Tianhua Zhou$^3$,
Xiaojia Chang$^3$,
Tongran Liu$^4$,} \\
\textbf{\ Jingbo Zhu$^{1,2}$,
Zhengtao Yu$^{5}$,
Tong Xiao$^{1,2}$\thanks{Corresponding author.}} \\
\ $^1$School of Computer Science and Engineering, Northeastern University, Shenyang, China \\
\ $^2$NiuTrans Research, Shenyang, China \\
\ $^3$Independent Researcher, Beijing, China \\
\ $^4$CAS Key Laboratory of Behavioral Science, Institute of Psychology, CAS, Beijing, China \\
\ $^5$Kunming University of Science and Technology \\
\ \texttt{\{wangchenglong, xiaotong\}@mail.neu.edu.cn}
}

\begin{document}

\maketitle
\lhead{arXiv preprint}

\begin{abstract}
Recent advances in reward modeling show a paradigm shift from discriminative reward models to generative reward models. However, despite their strong capabilities in response ranking, generative reward models have not realized their potential in reinforcement learning (RL). Our analysis reveals that this limitation arises from a \textit{mismatch} between the comparative nature of generative reward modeling and the scalar scoring paradigm adopted by existing RL algorithms. To bridge this gap, we propose a \textbf{\underline{R}}anking-based \textbf{\underline{R}}eward \textbf{\underline{C}}onstruction (RRC) approach, which enables generative reward models to provide more effective RL learning signals by deriving rewards from relative preference rankings. RRC introduces two complementary strategies: self-competitive ranking, which exploits comparisons among sampled responses, and anchor-guided ranking, which enables scalable ranking-based reward construction with a small set of reference responses. Experiments across open-ended chat and reasoning benchmarks demonstrate that RRC substantially improves RL training with generative reward models, achieving consistent gains over existing reward construction approaches (\textit{e.g.}, 35.8\%$\rightarrow$41.3\% on AlpacaEval2 and 8.0\%$\rightarrow$11.2\% on ArenaHardV2). Our code can be found at \url{https://github.com/wangclnlp/RRC}.
\end{abstract}

\section{Introduction}
% The rapid progress of large language models (LLMs) in recent years has been largely driven by the application of reinforcement learning (RL) \citep{bai2022training,wang2024hybrid,ji2025pku,xiao2025foundations}. A widely adopted approach is to train reward models to capture human preferences and subsequently fine-tune LLMs to align with these preferences. Reinforcement learning from human feedback (RLHF) is a representative example of this paradigm \citep{christiano2017deep,stiennon2020learning,ouyang2022training}. In such approaches, reward models are typically trained in a \textit{discriminative} manner to assign scalar reward scores to sampled responses, which serve as learning signals.

Reward models play a central role in reinforcement learning (RL) for large language models (LLMs) \citep{wang2024hybrid,ji2025pku,xiao2025foundations,wang2025worldpm,wang2026outcome}. They are typically trained in a \textit{discriminative} manner to assign scalar reward scores to sampled responses, which serve as learning signals for RL optimization.
Despite their empirical success, discriminative reward models suffer from inherent limitations. Building on the backbone of LLMs, discriminative reward models do not fully utilize the text generation capabilities of LLMs. Consequently, they do not take advantage of key intrinsic strengths of LLMs, such as instruction fine-tuning and explicit chain-of-thought (CoT) reasoning \citep{wei2022chain,zhang2024generative}.
This \emph{capability gap} is not merely a practical inconvenience but rather a fundamental barrier.
It prevents the reward model from capitalizing on insights revealed by recent advances in LLMs, including the use of large-scale  fine-tuning to enhance generalization \citep{shi2023specialist,yang2024unveiling} and the use of additional inference-time computation to improve performance \citep{muennighoff2025s1}.

To overcome these limitations, recent research has explored \textit{generative} approaches to developing reward models \citep{zhao2026genprm,hong2026think}.
Unlike traditional discriminative reward models, generative reward models represent preferences through text generation. For instance, given an input and a pair of responses, a generative reward model can produce a textual judgment that explicitly reflects which response is better and why. Although generative reward models have demonstrated strong performance in response ranking, their advantages have not yet been fully transferred to RL \citep{chen2025rm,wang2025gramv2,guo2025reward}. As a result, how to better leverage generative reward models within RL remains an open research question.
To date, existing strategies are significantly insufficient to address this challenge. For example, GenRM \citep{zhang2024generative} uses the probability of a designated preference token as the reward scores. However, this approach often suffers from probability collapse when CoT reasoning is employed. Additionally, GRAM \citep{wang2025gram} removes explicit CoT reasoning and generates only preference tokens, thereby producing scalar reward scores like discriminative reward models. 

All of the existing efforts overlook a fundamental distinction between discriminative and generative reward modeling paradigms: discriminative reward models are inherently designed to produce scalar reward scores, whereas generative reward models are naturally formulated to perform comparative ranking. We argue that forcing generative reward models into a scoring-based role within RL can lead to suboptimal preference prediction and policy optimization. To substantiate this claim, we design an analysis experiment that systematically compares discriminative and generative reward models across two settings (see Figure~\ref{fig:comparison_ranking_scoring}): response ranking and RL.
\textit{The results reveal a clear mismatch between the comparative nature of generative reward modeling and the scalar scoring paradigm adopted by existing RL algorithms.}

In this paper, we explore a solution to this limitation that leverages generative reward models solely for response ranking rather than requiring them to generate scalar reward scores in RL. To this end, we introduce a \textbf{\underline{R}}anking-based \textbf{\underline{R}}eward \textbf{\underline{C}}onstruction (RRC) approach, which constructs scalar reward scores directly from the relative preference orderings among sampled responses. Specifically, our RRC approach comprises two ranking mechanisms: self-competitive ranking and anchor-guided ranking, each providing a principled way to transform ranking information into usable reward scores. In self-competitive ranking, we can construct reward scores from the mutual comparisons among sampled responses. This basic idea is that responses generated by the current policy naturally form a competitive set, allowing relative quality differences to be exploited and guiding policy optimization. In anchor-guided ranking, we further consider scalability to larger sampled response sets by introducing a small number of anchor responses as reference points. This design reduces the number of calls to the generative reward model from $\mathcal{O}(m\cdot\log m)$ to $\mathcal{O}(m\cdot n)$, where $m$ is the number of sampled responses and $n$ is the number of anchor responses. The resulting reward scores can be seamlessly integrated into existing RL algorithms, such as GRPO \citep{shao2024deepseekmath} and DAPO \citep{yu2025dapo}.

Through extensive experiments on six benchmarks including open-ended chat and reasoning tasks, we show that RRC can provide a more effective way to leverage generative reward models within RL. Notably, compared to the probability-based reward baselines, RRC achieves significant improvements in RL performance (\textit{e.g.}, 35.8\%$\rightarrow$41.3\% on AlpacaEval2, 8.0\%$\rightarrow$11.2\% on the ArenaHardV2, and 52.9\%$\rightarrow$57.3\% on the MMLU-Redux). 

Beyond these gains, our results show that RL performance with RRC can be effectively scaled by increasing inference-time computation in generative reward models and by expanding the number of anchor responses (see Figure~\ref{fig:scaling_law_for_rrc}). This is appealing, as RRC introduces new dimensions for scaling RL. These empirical scaling behaviors are further supported by our theoretical analysis in Appendix~\ref{app:proof}, which suggests that RRC relaxes the constraints imposed by scalar rewards derived from discriminative reward models, thereby enabling a higher attainable performance ceiling for RL.

\section{Preliminaries}

\subsection{Training Reward Models}
We define a reward model as a function $r_{\phi}(x, y)$, where $\phi$  denotes the model parameters, $x$ is the input, and $y$ is the corresponding response. Throughout this work, an \textit{input} refers to an arbitrary token sequence provided to the LLM, such as ``\textit{Why the sky appears blue?}'', and a \textit{response} denotes the token sequence generated by the LLM conditioned on that input, such as ``\textit{This is because shorter-wavelength light is scattered more strongly by molecules}''. To date, mainstream reward model architectures can be categorized into two types: \textit{discriminative} and \textit{generative}.

\paragraph{Discriminative Reward Models.} 
Discriminative reward models produce scalar reward scores using a regression-style architecture. They are typically implemented as a stack of Transformer encoder layers, or as a Transformer decoder without a Softmax layer at the output. In practice, a pre-trained LLM is commonly used as the backbone. The concatenated input–response pair ($x,y$) is fed into the backbone model, and the final-layer hidden representation is then mapped to a scalar reward score via a linear projection. This model can be trained via a Bradley-Terry loss function \citep{bradley1952rank}:
\begin{equation}
\mathcal{L}_{\mathrm{d}} = - \mathbb{E}_{(x,y_{a},y_{b})\sim D_{r}} \big[ \log (\sigma (r_{\phi}(x,y_{a})-r_{\phi}(x,y_{b}))) \big]
\end{equation}
where $D_{r}$ is the training dataset consisting of tuples of input $x$ and response pair $(y_{a},y_{b})$ with the preference $y_{a} \succ y_{b}$. Once trained, the reward model can be naturally used as a scoring function that assigns a numerical reward $r_{\phi}(x,y)$.

% \begin{figure}[!t]
%     \centering
%     \vspace{4.5cm}
%     \caption{
%     Architecture of the Generative Reward Model.
%     The generative reward model leverages a pretrained LLM to predict a preference label from a given prompt directly. Optionally, it can incorporate explicit reward reasoning (e.g., CoT) before producing the final preference prediction.
%     }
%     \vspace{-2mm}
%     \label{fig:genertaive_rm_arch}
% \end{figure}

\paragraph{Generative Reward Models.} 
Compared to discriminative reward models, generative reward models fully leverage the text generation capabilities that LLMs are fundamentally designed for \citep{liang2025generative,wang2025gram}, producing reward signals through natural language generation rather than direct scalar scoring.
Specifically, they use an LLM to generate preference-related tokens, given a natural language prompt $c$ and a tuple $(x, y_a, y_b)$. The prompt describes the task in natural language, and the model predicts a label token $w$ that aligns with the human preference $l$, where $l = \text{A}$ denotes preference for $y_a$, and $l = \text{B}$ indicates preference for $y_b$. The model can be trained by
\begin{equation}
\mathcal{L}_{\mathrm{g}} = - \mathbb{E}_{(c,x,y_a,y_b,l) \sim D_r} \big[\log \pi_{\phi}(w=l|s)\big] \label{eq:gem-reward-modeling}
\end{equation}
where $s$ denotes the string $[c,x,y_a,y_b]$, and $\pi_{\phi}(\cdot)$ denotes the probability of token prediction.

Building on this generative formulation, recent studies further showed that framing reward prediction as a reasoning task can better exploit the reasoning capabilities of LLMs, leading to improved reward modeling performance \citep{chen2025rm,guo2025reward,wang2026mro}. In these approaches, the model is trained to generate explicit CoT,
\textit{i.e.}, analyzing and evaluating candidate responses individually, before producing the final preference prediction.
% as illustrated in Figure~\ref{fig:genertaive_rm_arch}.

\subsection{Applying Reward Models to RL}
In RL, the policy optimization objective can be written as $\max_{\theta}\mathbb{E}_{x \sim \mathcal{D}, o \sim \pi_{\theta}(\cdot | x)} [ r_{\phi}(x, o) ] $, where $x$ denotes an input prompt sampled from the dataset $\mathcal{D}$, $o$ denotes a response generated by the current policy $\pi_{\theta}(\cdot | x)$, and $\theta$ denotes the parameters of the policy model. 
A commonly used algorithm to optimize this objective is proximal policy optimization (PPO) \citep{schulman2017proximal}. However, PPO is often complex and unstable due to its reliance on an additional critic model to estimate advantages. To address this issue, GRPO \citep{shao2024deepseekmath} replaces the learned value function with relative rewards computed within a group of sampled responses $\{o_1, o_2, \ldots, o_m\}$, where $m$ denotes the group size. Specifically, the advantage for a sampled response $o_t$ is computed as $A_t = \frac{ r_{\phi}(x, o_t) - \mu }{ \sigma }$, where $\mu$ and $\sigma$ denote the mean and standard deviation of reward values within the sampled group, respectively.
Under this formulation, reward models are required to provide scalar-valued signals for each sampled response. Discriminative reward models naturally satisfy this requirement, as they are explicitly designed to assign scalar reward scores to individual responses. 

When applying generative reward models to RL, reward scores are typically constructed from the predicted probabilities of designated preference tokens \citep{zhang2024generative,wang2025gram}. Specifically, the reward score is computed for a single input–response pair ($x, o$). During the RL training, we first obtain a reference response $o_{\mathrm{ref}} = \arg\max_{o} \pi_{\theta}(o | x)$ via greedy decoding from the current policy. We then concatenate the context prompt $c$, the input $x$, the sampled response $o$, and the reference response $o_{\mathrm{ref}}$ into a single sequence $s_{o} = [c, x, o, o_{\mathrm{ref}}]$.
The generative reward model $\pi_{\phi}(\cdot)$ is prompted to predict which response is preferred, and the final reward for the pair ($ x, o $) is defined as the probability that $o$ is preferred over $o_{\mathrm{ref}}$. Concretely, if $o$ is designated as ``Response A'' in the prompt, the reward score is
\begin{equation}
r_{\phi}(x, o) = \pi_{\phi}(w=\text{A} \mid s_{o})
\label{eq:apply-generative-rm}
\end{equation}
where $w$ denotes the predicted preference token and the resulting reward score lies in the range $[0,1]$.

\section{Ranking vs. Scoring in Reward Models}

\begin{wrapfigure}{r}{0.52\textwidth}
    \centering
    \includegraphics[width=\linewidth]{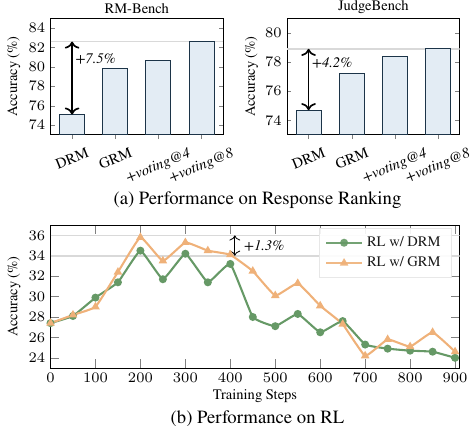}
    \caption{
    Comparison between discriminative and generative reward models under response ranking and RL settings. RL performance is evaluated using AlpacaEval2.
    }
    % \vspace{-4mm}
    \label{fig:comparison_ranking_scoring}
\end{wrapfigure}

Both discriminative and generative models are commonly adopted for reward modeling, but we found that generative models are better suited for response ranking, rather than for serving as a score-based role in RL. To study this issue, we trained both a discriminative and a generative reward model using the same labeled preference dataset with LLaMA-3.2-3B-Instruct serving as the backbone\footnote{The dataset includes human-annotated rationales, enabling explicit reward reasoning in the generative reward model.}.
We then evaluated these models on pairwise ranking benchmarks, including RM-Bench \citep{liu2024rm} and JudgeBench \citep{tan2024judgebench}, as well as in an RL setting where reward scores are constructed from the predicted probabilities of preference labels.
As shown in Figure~\ref{fig:comparison_ranking_scoring}, generative reward models significantly outperform discriminative ones on response ranking tasks. However, this advantage largely diminishes in the RL setting. Although preliminary, these results are consistent with prior findings reported by \cite{wang2025gram} and \cite{guo2025reward}: when trained on the same preference data, generative reward models significantly outperform discriminative models on response ranking, but yield only marginal improvements when used as a score-based role in RL.

We attribute this discrepancy to a mismatch between the optimization objective of generative reward models and the requirements of scalar reward scoring. Specifically, discriminative reward models are designed to directly predict a scalar reward score for each response, where the score represents the quality of the response and can be easily used as a learning signal in RL. In contrast, generative reward models are typically optimized to express preferences in the form of comparative judgments. 
In practice, when applying generative reward models in RL, we use the probability of predicted preference tokens as reward signals, thereby bridging textual preference prediction and scalar scoring. However, this suffers from inherent limitations, as illustrated in Figure~\ref{fig:main_image}(a). On one hand, when explicit reward reasoning is employed, preference token probabilities often collapse, making them poorly calibrated as reward scores. On the other hand, the predicted probabilities also reflect the model's confidence in its judgment, which is not necessarily aligned with human preference and might be noise. 
Therefore, the potential of generative reward models in preference modeling are weakened when they are directly used as a score-based role.

\section{Proposed Method}
We now describe our approach for better leveraging the comparative strengths of generative reward models in RL. We first briefly review the desiderata for reward construction in RL. We then introduce our RRC approach.
% which satisfies these desiderata by relying solely on response ranking.

\subsection{Desiderata for Reward Construction in RL}
Given an input $x$ and a sampled response $o$, an effective reward function $r(x, o)$ for LLM reinforcement learning should satisfy the following two properties.

\textbf{Property 1.} \textit{For any two responses ($o_i, o_j$) sampled under the same input $x$, the reward should preserve the preference ordering, \textit{e.g.}, $o_i \succ o_j \rightarrow r(x, o_i) > r(x, o_j)$.
}

We refer to this property as \textit{order preservation}. It ensures that policy updates are directionally consistent with preference judgments, so that responses more aligned with human preferences are consistently reinforced.

\textbf{Property 2.} \textit{The reward difference should reflect the relative quality gap between responses: $r(x, o_i) - r(x, o_j)  \propto  \Delta q_{ij}$, where $\Delta q_{ij}$ denotes the latent oracle preference margin between responses $o_i$ and $o_j$.}

We refer to this property as \textit{margin awareness}.
This property enables meaningful advantage estimates, preventing all preference differences from being treated as equally informative during optimization. As shown in Appendix~\ref{app:proof}, we further provide a proof from a policy gradient perspective, showing how this property directly affects the magnitude and effectiveness of optimization updates.

\begin{figure*}[!t]
    \centering
    \resizebox{\linewidth}{!}{
    \includegraphics[width=\linewidth]{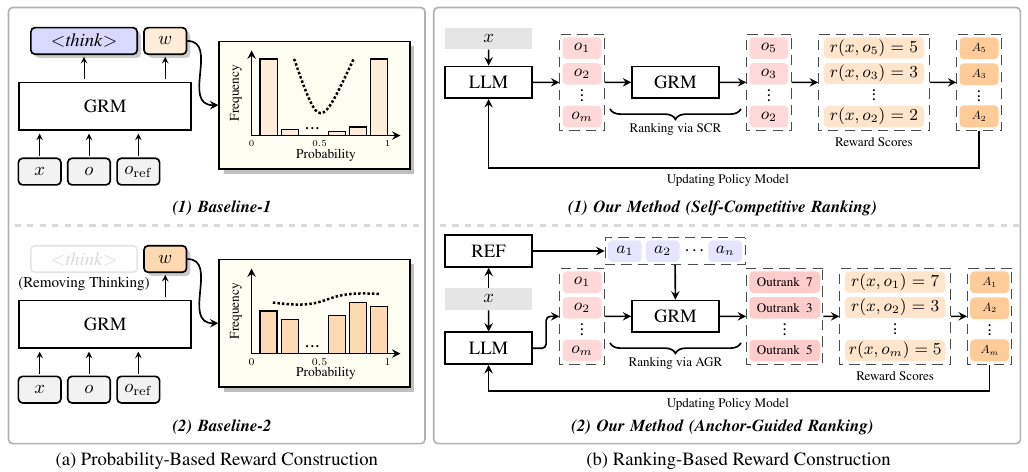}}
    \caption{
    \textbf{(a) Probability-Based Reward Construction.} We use the predicted probability of the preference token as the scalar reward signal (denoted as \textit{Baseline-1}). We also remove explicit reasoning to encourage output diversity and improve margin awareness (denoted as \textit{Baseline-2}). \textbf{(b) Ranking-Based Reward Construction.} We obtain reward scores from relative rankings obtained either among sampled responses or between sampled responses and anchor responses. Here, ``REF'' denotes the reference policy used to generate anchor responses. 
    The Python-style ranking algorithms for SCR (Alg.~\ref{alg:scr}) and AGR (Alg.~\ref{alg:agr}) are provided in the Appendix.
    }
    \label{fig:main_image}
\end{figure*}

\subsection{Ranking-Based Reward Construction}
When applying reward models in RL, optimization performance depends on whether the reward scores they provide satisfy both order preservation and margin awareness. Guided by these two properties, we redesign the reward construction process when using generative reward models. As illustrated in Figure~\ref{fig:main_image}(b), we propose an RRC approach. Unlike previous work, we do not utilize preference token probabilities to construct reward scores. Instead, we construct reward scores solely from response rankings. Specifically, RRC is implemented through two complementary ranking mechanisms: \textit{self-competitive ranking} and \textit{anchor-guided ranking}. 

\paragraph{Self-Competitive Ranking.}
The goal of self-competitive ranking (SCR) is to construct reward signals purely from the relative ordering among sampled responses, rather than relying on absolute scoring. In RL, responses generated under the same input naturally form a comparable set, and their relative quality provides a more reliable basis for reward construction than isolated scalar scores \citep{yuan2023rrhf,shao2024deepseekmath}. This motivates us to treat these responses as competitors in a self-contained ranking process. Formally, given an input $x$, we sample a group of responses $\{ o_1, o_2, \ldots, o_m \}$ from the current policy. We then query the generative reward model to perform pairwise preference judgments within this set. For each pair ($o_i, o_j$), the generative reward model predicts a preference relation:
\begin{equation}
    o_i \succ o_j \quad \text{or} \quad o_j \succ o_i
\end{equation}
These pairwise preferences induce a tournament graph over the response set. Based on the pairwise comparisons, we assign each response a reward score via the number of wins:
\begin{equation}
r(x,o_{i}) = \alpha \times \sum_{j \neq i} \mathbf{1}[o_i \succ o_j]
\end{equation}
where $\mathbf{1}(\cdot)$ denotes the indicator function and $\alpha$ denotes a scaling factor. Intuitively, responses that are preferred over more competitors receive higher reward scores, reflecting stronger relative quality within the sampled group.

This self-competitive construction satisfies both desiderata for reward construction in RL. Order preservation is guaranteed since higher-ranked responses receive larger rewards. Additionally, margin awareness emerges naturally, as reward differences reflect the number of relative wins against other responses, providing graded learning signals. 
% Importantly, since all comparisons are performed among responses generated by the current policy, the resulting rewards adapt dynamically as the policy evolves.

\paragraph{Majority Voting for Robust Ranking.}
Pairwise preference predictions from generative reward models can be stochastic, especially when the reward reasoning is involved, which may introduce noise into individual comparisons. To improve robustness, we allocate additional inference-time budget to stabilize preference estimation \citep{muennighoff2025s1,chang2025step}. 
Specifically, we adopt a majority voting scheme when aggregating pairwise preferences. For each response pair ($o_i, o_j$), we query the generative reward model multiple times under identical inputs and collect a set of preference predictions. The final preference relation is then determined by majority vote across these predictions.

\paragraph{Conflict-Aware Ranking Adjustment.}
Pairwise preference predictions produced by generative reward models are not guaranteed to be globally consistent. In practice, cyclic inconsistencies may arise, such as 
\begin{equation}
    o_i \succ o_j,\quad o_j \succ o_k,\quad \text{but}\quad o_k \succ o_i
\end{equation}
which violate transitivity and induce conflicts in the induced ranking. To mitigate this issue, we perform conflict-aware ranking adjustment by treating pairwise preferences as a directed weighted graph, as shown in Property 3.

\textbf{Property 3.} \textit{Given inconsistent pairwise preferences produced by a generative reward model, a Kemeny-rule-based aggregation can recover a globally consistent ranking that maximizes the weighted agreement with all pairwise preferences, where the weight of each preference is determined by its support degree (e.g., voting count) from the model.}

The proof of Property~3 is provided in Appendix~\ref{app:proof}. Specifically, given a sampled set of responses, we construct a directed weighted graph $G = (V, E)$, where each node $v_i \in V$ corresponds to a response $o_i$, and each directed edge $e_{ij} = (o_i \rightarrow o_j) \in E$ denotes a predicted preference $o_i \succ o_j$. The weight of each edge is defined as the number of times the preference $o_i \succ o_j$ is predicted by the generative reward model across multiple stochastic queries. Following the Kemeny rule’s core idea of minimizing global pairwise disagreements, we transform the conflict resolution problem into finding the minimum weight feedback edge set of $G$, where removing edges with the smallest total weight eliminates cycles while retaining as many high-weight preferences as possible. In practice, we adopt the efficient greedy heuristic from \cite{davenport2004computational} to iteratively select the pairwise preference with the largest weight difference between conflicting directions, fix its order according to the majority preference, and maintain the acyclicity of the graph via transitive closure propagation.

\paragraph{Anchor-Guided Ranking.}
\label{sec:agr}
While SCR provides a principled way to construct rewards from pairwise comparisons, it requires $\mathcal{O}(m\cdot\log m)$ preference queries for a group of $m$ sampled responses, which becomes computationally prohibitive when using generative reward models.
This quadratic cost limits scalability, especially in modern RL algorithms that rely on large-scale sampling to improve exploration and stability.
To solve this problem, we propose an anchor-guided ranking (AGR) mechanism that reduces the number of required comparisons while preserving relative ordering information. The basic idea is to introduce a small set of reference responses, referred to as \textit{anchors}, which serve as stable landmarks for estimating the relative quality of sampled responses. 

Formally, given an input $x$, we collect a small set of anchor responses $\{ a_1, a_2, \cdots, a_n \}$, where $n$ denotes the size of the anchor responses. Our experimental results show that even a small value of $n$ is sufficient to achieve strong performance (see Figure~\ref{fig:scaling_law_for_rrc}). For each sampled response $o_i$ and each anchor $a_k$, we query the generative reward model to predict a pairwise preference relation, indicating whether $o_i \succ a_k $ or $a_k \succ o_i$. Based on these comparisons, we define the reward score of the $i$-th sampled response as
\begin{equation}
r(x, o_{i}) = \alpha \times \sum_{k=1}^{n} \mathbf{1}[o_i \succ a_k]
\end{equation}
Intuitively, responses that outperform more anchors are considered to have higher relative quality.
Furthermore, our AGR naturally supports majority voting by querying the generative reward model multiple times for each anchor comparison $(o_i, a_k)$ and aggregating the preferences. This provides an additional inference-time scaling dimension to improve reward reliability, while still keeping the overall query complexity at $\mathcal{O}(m \cdot n)$.
In this way, we provide a comparison between SCR and AGR in Appendix~\ref{app:comparison_our_approaches}.
% Compared to SCR, AGR can reduce the number of generative reward model queries from $\mathcal{O}(m^2)$ to $\mathcal{O}(m \cdot n)$, enabling scalable reward construction in RL.

In this mechanism, anchor construction is key to its effectiveness. In this work, anchors are generated from a reference policy. As shown in Appendix~\ref{app:proof}, our theoretical analyses indicate that, compared to alternative designs such as selecting anchors from the current policy's samples, this strategy avoids coupling anchor quality with the evolving policy, thereby providing a more stable basis for comparison.

\section{Experiments}
We evaluated RRC in open-ended RL settings, where reward models rather than verifiable rules must provide rewards. 

\subsection{Experimental Setups}

\paragraph{Model Backbones.}
For reward model training, we employed LLaMA-3.1-8B-Instruct and LLaMA-3.2-3B-Instruct \citep{grattafiori2024llama} as backbone models. For RL training, we used LLaMA-3.1-8B-Instruct as the policy model. We also evaluated RRC on Qwen2.5-7B-Instruct \citep{team2024qwen2} to validate the effectiveness of RRC, as shown in Table~\ref{tab:results_on_qwen} in the Appendix.

\paragraph{Training Datasets.}
For reward model training, we used the HelpSteer3 \citep{wang2025helpsteer3}, which comprises 40.5K labeled preference examples. Each example includes human-written feedback and comparative analysis, which we treated as rationales when training generative reward models. For discriminative reward models, we used the same binary preference labels (\textit{i.e.}, which response is preferred). Still, we did not provide access to the textual rationales to ensure a controlled and fair comparison between the two reward modeling approaches. For RL training, we followed the same supervised fine-tuning (SFT) and RL datasets used in \cite{bhaskar2025language}. The SFT dataset consisted of 6K examples, while the RL dataset contained 7.5K examples. 
These datasets were designed for open-ended RL and covered a diverse set of tasks, reflecting realistic instruction-following scenarios.

% sampling 8 (left) 
% sampling 16 (right)

% (a) 1,2,4,8,16,32,64,128
% ranking-3B ranking-8B

\paragraph{Settings.}
We trained both discriminative and generative reward model baselines for one epoch using a learning rate of 1e-5 and a batch size of 256. For all RL training runs in our main experiments, we used the GRPO algorithm, with a learning rate of 1e-6 and a batch size of 128, while keeping all other hyperparameters at their default values. To evaluate the robustness of our RRC, we conducted RL training under two policy settings. In the first setting, the policy model generated explicit CoT reasoning before producing the final response (denoted as \textit{reasoning-augmented policy}). In the second setting, the policy model generated only the final response without intermediate CoT reasoning. In the former setting, the generated reasoning paths were not provided to the reward models; only the final responses were evaluated. Additional training details are provided in Appendix~\ref{app:experiments}.

\paragraph{Baselines.}
Our main baseline was the probability-based reward construction approach (denoted as \textit{PRC}), where the predicted probability of the preference token was used as the scalar reward score \citep{zhang2024generative,wang2025gramv2}. Additionally, we considered a variant that removed explicit reward reasoning to encourage output diversity and improve margin awareness \citep{wang2025gram} (denoted as \textit{PRC+Removing Reasoning}). We also included a discriminative reward model trained on the same preference dataset as the generative reward models. Furthermore, we compared RRC with several offline preference optimization approaches, including \textit{DPO} \citep{rafailov2023direct} and \textit{SimPO} \citep{meng2024simpo}, trained on the same preference data.
% , enabling a controlled comparison between discriminative and generative reward modeling paradigms.

\begin{table*}[!t]
    \centering
    \resizebox{0.99\linewidth}{!}{

% \usepackage{multirow}
\begin{tabular}{lcrcccccccc}
\toprule[1.1pt]
\multirow{2}{*}{Method} & \multicolumn{5}{c}{\textit{w/ Thinking}}      & \multicolumn{5}{c}{\textit{w/o Thinking}}  \\  \cmidrule(l){2-6} \cmidrule(l){7-11}
& AE2 & AH2 & WiB & MM$_{\text{R}}$ & MATH & AE2  & AH2 & WiB & MM$_{\text{R}}$ & MATH \\ \midrule
SFT (\textit{w/ LLaMA-3.1-8B-Instruct}) & 27.4 & 6.4 & 46.6 & 55.4 & 38.8  & 23.7  &6.2  &44.5 & 50.4 &  36.4                          \\
\quad + DPO    &29.8  &7.0    & 48.8   &52.1  & 39.6 & 26.1  & 7.5 & 45.3  & 51.4 & 36.8\\
\quad + SimPO &31.2  &7.2  & 50.3 &54.7 & 40.4 & 28.2 & 7.0  & 48.4 & 50.2  & 38.6\\ \midrule
\multicolumn{7}{l}{\textbf{\textit{RL Training with 3B-Scale Reward Models}}}   \\  \midrule
\quad + DRM         &30.5  &7.0    &48.7  & 52.3  & 42.6 & 27.6  &7.2  &46.3 & 50.2  &40.4 \\
\quad + GRM w/ PRC  &31.4  &7.8    & 50.2  & 52.4  & 44.8  &28.4 &6.8  &48.1 & 51.7  &41.2 \\
\qquad + \textit{Removing Thinking}  &32.4  &7.2  & 51.7  &53.6  &43.0 &28.8 &8.0  &49.2  & 50.0 & 41.2
\\  \hdashline
\quad + GRM w/ RRC-SCR  &36.4  &8.8   & 55.6  &54.4 &45.6 &33.1 &8.2 & 53.0 &  52.4 &42.0 \\ 
\qquad + \textit{voting@8}  &\textbf{37.8}  &9.8  & 56.7 &\bf56.7 &47.2 &33.8 &9.0  &53.2  & 53.7 & 44.8\\
\quad + GRM w/ RRC-AGR    &35.8  &9.8   & 56.8  &55.0 &46.4 &32.8 &8.8  &52.9  &  52.7 &45.4 \\ 
\qquad + \textit{voting@8}  &37.5  &\textbf{10.2} & \bf58.2  &56.4 &\bf48.4 &\bf34.9  & \bf9.4  &\bf54.7  & \bf 54.5 & \bf45.8\\
\midrule
\multicolumn{7}{l}{\textbf{\textit{RL Training with 8B-Scale Reward Models}}}   \\ \midrule
\quad + DRM                          & 33.2  &7.4 & 49.8 &52.7 & 42.0 & 30.1 &7.0  &47.2 & 51.4 & 40.2 \\
\quad + GRM w/ PRC                   & 35.8  &7.8 & 48.3 &51.4 & 41.2 &32.3  & 7.2 & 47.5  & 50.6 & 38.8 \\
\qquad + \textit{Removing Thinking}  &35.1  &8.0 & 50.1  &52.9 &43.6 & 32.5  & 7.4  & 46.5  & 50.8 & 41.6 \\ \hdashline
\quad + GRM w/ RRC-SCR           &38.7  & 9.0 & 56.6 &55.1 &46.4 &34.6  &8.8  &53.3 & 53.6 & 44.8\\ 
\qquad + \textit{voting@8}       &40.0  & 10.4 & 57.8  &\bf57.3 &47.4 &34.8 &8.2  & 54.5 & 54.8 & 45.2\\
\quad + GRM w/ RRC-AGR           &39.4  & 11.0 &58.1   &55.4 &47.8 &\bf35.2 & 9.0 & 55.1 &  53.4 & 45.0\\ 
\qquad + \textit{voting@8}       &\textbf{41.3}  & \textbf{11.2} & \bf59.8  &56.9  &\bf48.6 &34.8 &\bf9.8  &\bf55.8 &\bf 56.2 & \bf46.8 \\
\bottomrule[1.1pt]
\end{tabular}
    }
    \caption{
    Performance of models fine-tuned with explicit reasoning (\textit{w/ Thinking}) and without explicit reasoning (\textit{w/o Thinking}). Best results within each group are shown in \textbf{bold}. A dotted line separates baselines from our RRC. \textit{+voting@8} is that we use a majority voting of 8 to enhance RRC, together with conflict-aware ranking adjustment in the SCR. 
    3B-scale and 8B-scale reward models denote that models are trained based on LLaMA-3.2-3B-Instruct and LLaMA-3.1-8B-Instruct, respectively.
    DRM: Discriminative Reward Model; GRM: Generative Reward Model; PRC: Probability-Based Reward Construction.
    }
    \label{tab:main_results}
\end{table*}

\paragraph{Evaluation.}
We focused on three open-ended chat benchmarks: AlpacaEval2 \citep{alpaca_eval}, ArenaHardV2 \citep{li2024crowdsourced}, and WildBench \citep{lin2024wildbench}, which are commonly used to assess the performance of LLMs trained with open-ended RL. We also evaluated our methods on knowledge and math reasoning benchmarks, including MMLU-Redux \citep{gema2025we} and MATH-500 \citep{lightman2023let}.

\subsection{Main Results}
Table~1 reports the performance of models fine-tuned from the LLaMA-3.1-8B-Instruct model. First, comparing DRM and GRM w/ PRC, we observe trends consistent with those in Figure~\ref{fig:comparison_ranking_scoring}: using probability-based reward construction does not effectively translate the ranking advantage of generative reward models into RL. For example, under 3B-scale reward models with thinking, GRM w/ PRC achieves 31.4\% on AlpacaEval2 and 7.8\% on ArenaHardV2, only marginally better than DRM (30.5\%/7.0\%). Similar gaps are observed in the non-thinking setting (28.4\% vs. 27.6\% on AlpacaEval2). Second, across both 3B- and 8B-scale reward models, our proposed RRC approaches consistently outperform all baselines by a clear margin. The results show that constructing rewards from rankings rather than token probabilities is substantially more effective for generative reward models in RL. Third, we observe that the proposed majority voting mechanism further boosts performance. For example, under 3B reward models with thinking, RRC-SCR improves from 36.4\% to 37.8\% on AlpacaEval2 after applying \textit{voting@8}. This suggests that aggregating multiple stochastic preference judgments helps reduce noise in pairwise comparisons and yields more robust rankings.

Furthermore, we further investigate the performance gain on two different ranking approaches in RRC. From the results, we find that AGR generally achieves stronger or comparable performance, especially when combined with voting. We attribute this advantage to the use of anchors generated from a fixed reference policy, which provides more stable and consistent comparison baselines and avoids coupling reward estimation to the rapidly evolving policy distribution.

\section{Analysis}
\subsection{Scaling Behavior of RRC}
\label{sec:scaling_law_rrc}
\begin{figure*}[!t]
    \centering
    \resizebox{\linewidth}{!}{
     \includegraphics[width=\linewidth]{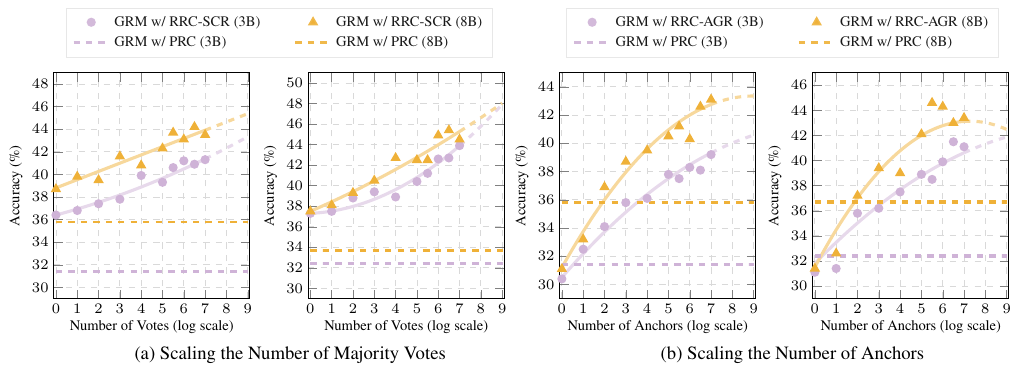}}
    \caption{
    Scaling behavior of RRC.
    }
    \label{fig:scaling_law_for_rrc}
\end{figure*}

We investigate the scaling behavior of RRC along two key dimensions: 1) the number of votes used by the generative reward model, and 2) the number of anchors employed in AGR. 
All experiments are conducted using 3B-scale reward models, under two different sampling settings (sampling 8 and sampling 16 responses per prompt). The results are summarized in Figure~\ref{fig:scaling_law_for_rrc}.
From the results, we see a clear and consistent scaling trend with respect to both the number of votes and the number of anchors. 
Specifically, increasing the number of votes leads to monotonic improvements in accuracy, indicating that more reliable preference aggregation from the generative reward model yields better-aligned reward signals and thus more effective policy optimization. 
Similarly, increasing the number of anchors in AGR consistently improves performance, suggesting that denser anchor comparisons provide finer-grained relative quality estimates, which translate into more informative advantage signals in RL.
Notably, the performance gains exhibit diminishing returns as the number of votes or anchors continues to increase, which is characteristic of scaling-law behavior \citep{kaplan2020scaling}. 
This shows that while larger comparison budgets improve reward calibration and margin estimation, the marginal benefit gradually decreases beyond a certain scale, indicating a trade-off between computational cost and performance. 

From the scaling behavior of the AGR, we can see that a relatively small number of anchors (\textit{e.g.}, 8 and 16) is already sufficient to yield substantial performance gains. 
Interestingly, when the number of anchors becomes very large (\textit{e.g.}, 256), performance tends to saturate or even slightly degrade. 
We hypothesize that under large-scale anchor sampling, it becomes increasingly difficult to maintain sufficient diversity among anchor responses, leading to redundant or low-informative comparisons. 

\section{Related Work}

\paragraph{Reward Modeling.}
Reward models play a central role in RLHF and other alignment strategies, such as direct preference optimization (DPO) and rejection sampling \citep{lee2021discriminative,rafailov2023direct,wang2024esrl,zhou2024prior}. 
Recent works on improving reward models could be classified into two groups. The first group focused on addressing the scarcity of preference data. For example, several studies constructed large-scale training datasets, either in task-specific settings \citep{stiennon2020learning,xu2024contrastive} or as general-purpose preference datasets \citep{cui2023ultrafeedback}. Additionally, GRAM \citep{wang2025gram} and POLAR \citep{dou2025pre} explored the use of unlabeled data for pre-training reward models. The second group explored stronger reward modeling approaches \citep{coste2023reward,min2024dynamic}. A major development in this line of work was the transition from discriminative to generative reward modeling. Building on this shift, recent studies showed that incorporating explicit reward reasoning (\textit{i.e.}, CoT) into generative reward models was crucial for improving reward modeling \citep{chen2025rm,wang2025gramv2,chen2025rm,guo2025reward,wang2025unified-understanding,wang2026unified}. 
% However, despite their strong performance in response ranking, these advantages have not fully translated into more effective learning signals for RL.
% Furthermore, \cite{chen2025rm} and \cite{guo2025reward} leveraged large-scale RL to improve the reward reasoning capabilities.

\paragraph{RL with Reward Models.}
% RL has become a standard post-training paradigm for LLMs, with RLHF being the most representative framework \citep{ouyang2022training,bai2022training}. 
% In RLHF, reward signals play a central role, and their quality directly affects the performance of the resulting models. 
In RLHF, one straightforward way to construct reward signals is to design task-specific rules, an approach commonly referred to as reinforcement learning with verifiable rewards (RLVR), which has been widely used to improve the reasoning capabilities of LLMs \citep{guo2025deepseek,mu2025dissecting,mroueh2025reinforcement}. While such rule-based rewards can provide accurate supervision, their applicability is typically limited to narrowly defined tasks and does not generalize well to open-ended, real-world tasks \citep{bhaskar2025language,liang2026dual}. As a result, the most widely adopted approach is to train reward models to provide learning signals \citep{ouyang2022training,wang2024hybrid}. In practice, this is typically implemented by training discriminative reward models that assign scalar scores to sampled responses. Motivated by the recent success of generative reward models, more recent studies have also explored using generative models in this score-based role \citep{zhang2024generative,wang2025gram}. Different from these approaches, we argue that treating generative reward models solely as scalar score predictors underutilizes their comparative modeling capabilities, which motivates us to propose RRC. Recent efforts, such as \cite{song2025reward}, have also investigated reference-based reward construction for generative reward models. However, these approaches still focus on deriving scalar rewards from reference comparisons, whereas RRC explores ranking-based reward construction.

% , \textit{e.g.}, treating the probability of preference tokens as reward scores \citep{zhang2024generative,wang2025gram}. 
% However, these approaches overlook the mismatch between the inherent comparative strengths of generative reward models and the score-based role imposed by RL.

\section{Conclusions}
We proposed an RRC approach for using generative reward models with reward reasoning capabilities in RL. 
Our approach fully exploits the inherent strength of generative reward models in response ranking, rather than requiring them to produce scalar scores. 
We conducted extensive experiments on six benchmarks, and the results show that RRC enables more accurate and reliable reward signals.

\bibliography{open_ended_rl}
\bibliographystyle{iclr_template/iclr2027_conference}

\appendix

\clearpage

% \begin{center}
% \makebox[\textwidth][c]{%
% \begin{minipage}{1.0\textwidth}
% \centering
% \textbf{\Large RRC: Unlocking Generative Reward Models in LLM Reinforcement Learning
% via Ranking-Based Reward Construction} \\
% \large Supplementary Material
% \end{minipage}}
% \end{center}

\vspace{1mm}
\section{Proofs for Theoretical Results}
\label{app:proof}
In this section, we present the proofs of three theoretical results. The first result establishes the importance of margin awareness in reward construction for RL. The second result shows that Kemeny-rule-based aggregation can recover a globally consistent ranking that maximizes the weighted agreement with all pairwise preferences. The third result demonstrates the advantage of using a reference policy to generate anchor responses.

\textbf{Property 1.}  \textit{The reward difference should reflect the relative quality gap between responses: $r(x, o_i) - r(x, o_j)  \propto  \Delta q_{ij}$, where $\Delta q_{ij}$ denotes the latent oracle preference margin between responses $o_i$ and $o_j$.}

\textbf{Proof:} For any two responses $o_i$ and $o_j$ sampled under the input $x$, the reward gap should reflect the underlying quality gap:
\begin{align}
    r(x,o_i)-r(x,o_j)\ \propto\ \Delta q_{ij},
    \qquad
    \Delta q_{ij}\triangleq q(x,o_i)-q(x,o_j)
\end{align}
where $q(x,o)$ denotes the latent (oracle) preference quality, and $\Delta q_{ij}$ is the latent oracle preference margin. 
Assume there exists an oracle quality function $q(x,o)$ that characterizes the true preference intensity for response $o$ given input $x$. The ideal RL objective is
\begin{align}
J_q(\theta)\ \triangleq\ \mathbb{E}_{x\sim\mathcal{D},, o\sim\pi_\theta(\cdot|x)}\big[q(x,o)\big]
\end{align}
By the policy gradient theorem, its gradient can be written as
\begin{align}
\nabla_\theta J_q(\theta) = 
\mathbb{E}_{x,o}\Big[q(x,o)\nabla_\theta \log \pi_\theta(o|x)\Big]
\end{align}
In practice, GRPO-style RL algorithms typically optimize a surrogate objective using a \textit{group} of sampled responses $\{o_1,\cdots,o_m\}$ for the same $x$, and define a normalized group-relative advantage:
\begin{align}
A_i^{(r)} = 
\frac{r_i-\mu_r}{\sigma_r},
\qquad
r_i\triangleq r(x,o_i),
\quad
\mu_r\triangleq \frac{1}{m}\sum_{k=1}^m r_k,
\quad
\sigma_r^2\triangleq \frac{1}{m}\sum_{k=1}^m (r_k-\mu_r)^2
\end{align}
Ignoring clipping and other stabilizers for clarity, the corresponding RL update direction is
\begin{align}
g_r(\theta)
\ \triangleq
\mathbb{E}_{x}\left[\sum_{i=1}^m A_i^{(r)} \nabla_\theta \log \pi_\theta(o_i|x)\right]
\end{align}
Hence, the relative magnitudes of policy updates across sampled responses are governed by the normalized reward gaps $A_i^{(r)}$. This demonstrates that not only the ordering of rewards, but also the proper calibration of reward differences, is crucial for effective optimization.  

Furthermore, assume that for any fixed $x$, reward scores from reward models are an affine transformation of oracle qualities:
\begin{align}
r(x,o) = bq(x,o) + d,\qquad b>0
\end{align}
Then the advantage computed from $r$ coincides with the advantage computed from $q$:
\begin{align}
A_i^{(r)} = A_i^{(q)} \ \triangleq\ \frac{q_i-\mu_q}{\sigma_q},
\quad
q_i\triangleq q(x,o_i),
\quad
\mu_q\triangleq \frac{1}{m}\sum_{k=1}^m q_k,
\quad
\sigma_q^2\triangleq \frac{1}{m}\sum_{k=1}^m (q_k-\mu_q)^2
\end{align}
Under the affine assumption,
\begin{align}
\mu_r = \frac{1}{m}\sum_{k=1}^m (b q_k+b) = b\mu_q + d
\end{align}
and
\begin{align}
\sigma_r^2 = \frac{1}{m}\sum_{k=1}^m \big(b q_k + d - (b\mu_q+d)\big)^2
= \frac{1}{m}\sum_{k=1}^m (b(q_k-\mu_q))^2
= b^2\sigma_q^2
\end{align}
so $\sigma_r = b\sigma_q$ since $b>0$. Therefore,
\begin{align}
A_i^{(r)} = 
\frac{b q_i+b-(b\mu_q+d)}{b\sigma_q} =
\frac{q_i-\mu_q}{\sigma_q}=A_i^{(q)}
\end{align}
Based on this equation, the RL update direction induced by $r$ becomes
\begin{align}
g_r(\theta) = 
\mathbb{E}_{x}\left[\sum_{i=1}^m A_i^{(q)} \nabla_\theta \log \pi_\theta(o_i|x)\right]
\end{align}
which is precisely the group-normalized counterpart of the oracle policy gradient. In particular, the \textit{relative weighting} assigned to different sampled responses is governed by ($q_i-\mu_q$), \textit{i.e.}, the oracle quality margins within the sampled group. Hence, to maintain oracle-consistent update geometry, reward differences must scale proportionally with oracle differences:
\begin{align}
r_i-r_j = b(q_i-q_j)= b \Delta q_{ij}
\end{align}
This is exactly the desired margin-awareness property:
\begin{align}
r(x,o_i)-r(x,o_j)\ \propto\ \Delta q_{ij}
\end{align}

Building on the above proof, we further analyze how our SCR and AGR mechanisms satisfy the property of margin awareness.
In SCR, reward scores are constructed from pairwise comparisons among sampled responses:
\begin{align}
r(x,o_i) = \alpha \sum_{j\neq i} \mathbf{1}[o_i \succ o_j]
\end{align}
Thus, the reward difference between the two responses satisfies
\begin{align}
r_i - r_j
= \alpha \sum_{k} \big(\mathbf{1}[o_i \succ o_k] - \mathbf{1}[o_j \succ o_k]\big)
\end{align}
This difference equals the net number of wins of $o_i$ over the rest of the group relative to $o_j$. Intuitively, if $o_i$ is substantially better than $o_j$, it will defeat more competitors and lose to fewer, yielding a larger reward gap. Conversely, if two responses are close in quality, their win counts will also be similar. 

In AGR, rewards are constructed via comparisons with a fixed set of anchor responses:
\begin{align}
r(x,o_i) = \alpha \sum_{k=1}^{n} \mathbf{1}[o_i \succ a_k]
\end{align}
where anchors ${a_k}$ are generated from a reference policy.
Here, the reward difference becomes
\begin{align}
r_i - r_j
= \alpha \sum_{k=1}^{n} \big(\mathbf{1}[o_i \succ a_k] - \mathbf{1}[o_j \succ a_k]\big)
\end{align}
which measures how much better $o_i$ performs than $o_j$ relative to the same quality benchmarks. If $o_i$ consistently outperforms anchors that $o_j$ fails against, the reward gap increases accordingly. This mechanism effectively estimates the relative position of each response along a shared quality scale defined by the anchors. Thus, reward differences reflect how much higher one response ranks than another with respect to common reference points, providing a stable and scalable approximation of oracle margin differences.

This completes the proof of this property.

\textbf{Property 2.}  \textit{Given inconsistent pairwise preferences produced by a generative reward model, a Kemeny-rule-based aggregation can recover a globally consistent ranking that maximizes the weighted agreement with all pairwise preferences, where the weight of each preference is determined by its support degree (\textit{e.g.}, voting count) from the model.}

\textbf{Proof:} Let $O=\{o_1,\ldots,o_m\}$ be the response set. Suppose the generative reward model outputs (possibly inconsistent) pairwise preferences with nonnegative weights ${w_{ij}}_{i\neq j}$, where $w_{ij}$ represents the support for the preference $o_i \succ o_j$ (\textit{e.g.}, the number of votes in favor of $o_i \succ o_j)$. This defines a weighted directed graph $G=(V,E)$ with $V=O$ and directed edges $i\to j$ of weight $w_{ij}$. A globally consistent ranking is a total order $\pi$ over $O$. Any total order $\pi$ induces a transitive tournament orientation: we say $\pi$ agrees with an edge $i\to j$ if $i$ is ranked ahead of $j$ under $\pi$ (denoted $i \prec_{\pi} j$ or equivalently $o_i \succ_{\pi} o_j$).
Define the weighted agreement of $\pi$ with the pairwise preferences as
\begin{align}
\mathrm{Agree}(\pi) :=
\sum_{i \prec_{\pi} j} w_{ij}
\end{align}
\textit{i.e.}, the total weight of directed preferences that are consistent with the total order $\pi$. Equivalently, define the weighted disagreement (\textit{a.k.a}, feedback weight) as
\begin{align}
\mathrm{Disagree}(\pi)
:=
\sum_{i \prec_{\pi} j} w_{ji}
\end{align}
which corresponds to the total weight of edges that point against the ordering induced by $\pi$.
Note that for any unordered pair ${i, j}$, exactly one of $i \prec_{\pi} j$ or $j \prec_{\pi} i$ must hold.
Hence,
\begin{align}
\mathrm{Agree}(\pi)+\mathrm{Disagree}(\pi) =
\sum_{i<j}\big(w_{ij}+w_{ji}\big) =: C
\end{align}
where $C$ is a constant independent of $\pi$. Therefore,
\begin{align}
\arg\max_{\pi}\mathrm{Agree}(\pi) =
\arg\min_{\pi}\mathrm{Disagree}(\pi)
\end{align}
This is precisely the Kemeny optimal aggregation objective in the weighted setting: find the total order $\pi^*$ that minimizes the total weight of pairwise disagreements \citep{simjour2009improved,betzler2014theoretical}.
Finally, because $\pi^*$ is a total order, it is globally consistent by construction: if $i \prec_{\pi^*} j$ and $j \prec_{\pi^*} k$, then necessarily $i \prec_{\pi^*} k$. Thus, the Kemeny-rule solution $\pi^*$ yields a globally consistent ranking that maximizes weighted agreement with the pairwise preferences, with weights determined by support degrees (\textit{e.g.}, voting counts).

This completes the proof of this property. 

\textbf{Property 3.} \textit{Compared to selecting anchors from samples of the current policy $\pi_{\theta}(\cdot)$, using a reference policy $\pi_{\mathrm{ref}}(\cdot)$ to generate anchor responses decouples anchor quality from policy updates. This mitigates feedback loops between policy improvement and anchor quality, reducing distributional drift and yielding more stable reward signals during training.}

\textbf{Proof:} In anchor-guided ranking (see Section~\ref{sec:agr}), the reward for a sampled response $o_i \sim \pi_\theta(\cdot|x)$ is constructed as
\begin{align}
r_\theta(x, o_i)
= \alpha \sum_{k=1}^n \mathbf{1}[o_i \succ a_k ,],
\quad a_k \sim q_\theta(\cdot|x)
\end{align}
where $q_\theta$ denotes the anchor-generating distribution. Here, we consider two strategies for generating anchors: 1) On-policy anchors: $q_\theta(\cdot|x) = \pi_\theta(\cdot|x)$ and 2) Reference anchors: $q_\theta(\cdot|x) = \pi_{\mathrm{ref}}(\cdot|x)$, independent of $\theta$. 

Here, we call the gradient of the RL optimization objective \textit{w.r.t} $\theta$:
\begin{align}
\nabla_\theta J(\theta) = 
\underbrace{
\mathbb{E}_{x,o}\left[
r_\theta(x,o)\nabla_\theta \log \pi_\theta(o|x)
\right]
}_{\text{policy improvement term}}
+
\underbrace{
\mathbb{E}_{x,o}\left[
\nabla_\theta r_\theta(x,o)
\right]
}_{\text{baseline drift term}}
\end{align}
The second term is non-zero because anchors are also drawn from $\pi_\theta$:
\begin{align}
\nabla_\theta r_\theta(x,o) = \alpha \sum_{k=1}^n
\mathbb E_{a_k \sim \pi_\theta}
\big[
\mathbf{1}[o \succ a_k]\nabla_\theta \log \pi_\theta(a_k|x)
\big]
\end{align}

Therefore, each policy update simultaneously: 1) shifts probability mass toward better responses $o$, and 2) improves the anchor set used as the comparison baseline. As a result, the reward assigned to a fixed response $o$ may decrease even if its absolute quality does not, simply because anchors become stronger. Hence, the optimization target itself changes with $\theta$, forming a moving-target objective:
\begin{align}
\theta_t  \Rightarrow  \pi_{\theta_t}  \Rightarrow  \text{stronger anchors}
 \Rightarrow  \text{shifted rewards}
\Rightarrow \nabla_\theta J(\theta_t) \text{changes}
\end{align}
This violates the stationarity assumption implicitly required by stochastic gradient ascent and increases both gradient variance and training instability. 

When anchors are sampled from a fixed reference policy,
\begin{align}
r(x,o) = \alpha \sum_{k=1}^n \mathbf{1}[o \succ a_k],\quad a_k \sim \pi_{\mathrm{ref}}
\end{align}
which is independent of $\theta$. Thus,
\begin{align}
\nabla_\theta r(x,o) = 0
\end{align}
and the gradient reduces to
\begin{align}
\nabla_\theta J(\theta) = 
\mathbb{E}\left[
r(x,o)\nabla_\theta \log \pi_\theta(o|x)
\right]
\end{align}
which corresponds to optimizing a fixed reward landscape. In this case, policy updates only affect the sampling distribution of candidate responses, while the evaluation baseline remains stationary. Hence, the optimization objective does not drift across iterations, satisfying the standard assumptions of stochastic optimization and yielding more stable training dynamics.

This completes the proof of this property.

\section{Additional Experimental Details and Results}
\label{app:experiments}
In this section, we provide additional experimental details and present the experimental results of our RRC approach.

\begin{table}[!t]
    \centering
    \resizebox{\linewidth}{!}{
    
\begin{tabular}{lcccccccccc}
\toprule[1.1pt]
\multirow{2}{*}{Method} & \multicolumn{5}{c}{\textit{w/ Thinking}}      & \multicolumn{5}{c}{\textit{w/o Thinking}}  \\  \cmidrule(l){2-6} \cmidrule(l){7-11}
& AE2 & AH2 & WiB & MM$_{\text{R}}$ & MATH & AE2  & AH2 & WiB & MM$_{\text{R}}$ & MATH \\ \midrule
SFT (\textit{w/ Qwen2.5-7B-Instruct}) & 27.1 &5.6 &44.6 &60.3 & 45.6 & 25.6 &5.2 & 46.2 &62.7 &47.8          \\
\quad + DPO & 29.6  & 6.2   & 48.4 & 58.4 & 42.4  & 28.7 &7.6 &47.8 &60.4 &45.6   \\
\quad + SimPO &30.4  & 7.0  & 50.3 & 56.7 & 45.8  & 29.1 &7.8 &48.9 &58.2 &46.2 \\ \midrule
\multicolumn{7}{l}{\textbf{\textit{RL Training via 3B-Scale Reward Models}}}   \\  \midrule
\quad + DRM  & 27.8  &6.8  &46.6  & 58.2 & 46.8 & 26.8 &7.4 &46.0 &59.1 &48.8  \\
\quad + GRM w/ PRC  &30.9  &7.2   & 48.3 & 56.4 & 48.8 & 28.1 &7.2 &47.8 &60.4 &48.0  \\
\qquad + \textit{Removing Thinking}  &28.4  &6.4 &46.6  & 55.3 & 45.2 & 27.6 &6.8 &46.8 &61.7 &48.4  \\  \hdashline
\quad + GRM w/ RRC-SCR   &33.2  & 7.8 & 51.2  &58.4 & 48.0 & \bf34.8 &7.6 &48.6 &62.2 &49.8  \\ 
\qquad + \textit{voting@8}  & 35.6 & 8.4 & \bf52.8 &60.5 &50.0 &35.7 &8.2 &49.2 &64.7 &48.6 \\
\quad + GRM w/ RRC-AGR   & 34.5 & 8.6  & 52.8 & 62.6 & 51.8 & 32.8 & 8.0 & 49.6 & 65.7  & 51.4  \\ 
\qquad + \textit{voting@8}  & \bf35.8 & \bf9.4  & 52.4 & \bf63.4 & \bf52.6 & 33.8 & \bf8.6 & \bf50.6 & \bf 67.6 & \bf52.8 \\
\bottomrule[1.1pt]
\end{tabular}
    } 
    \caption{
    Performance of models fine-tuned from Qwen2.5-7B-Instruct model. All experiments are conducted using 3B-scale reward models.
    }
    \label{tab:results_on_qwen}
\end{table}

\subsection{Experimental Details}

\paragraph{Training Setups.}
For the experiments reported in Figure~\ref{fig:comparison_ranking_scoring} and Table~\ref{tab:main_results}, we set the group sampling size to $m=8$ and the number of anchor responses to $n=8$. We also evaluated other group and anchor sizes in Figure~\ref{fig:scaling_law_for_rrc} to study the impact of these hyperparameters. For RRC, we set the scaling factor $\alpha$ to 0.1. Following \cite{bhaskar2025language}, we additionally incorporated a format reward to encourage structured outputs in the reasoning-augmented setting. The final reward is computed as a combination of the ranking-based reward and the format reward:
\begin{equation}
    r(o,x) = r_{\phi}(o,x) + r_{\mathrm{format}}(o)
\end{equation}
Here, we define the format reward $r_{\mathrm{format}}(\cdot)$ as a binary indicator: it equals 1 if the sampled response follows the required structure with explicit \texttt{<think>} and \texttt{<answer>} tags, and 0 otherwise. 
When querying the generative reward model, we account for the positional bias reported in \cite{wang2024large} by randomly assigning responses to the ``Response A'' and ``Response B'' slots in each pairwise comparison.
Additionally, during RL training, we saved a checkpoint every 100 steps. We then selected the best checkpoint based on performance on a validation set. The validation set was sampled from the training distribution and consists of 500 examples. For validation, we used \texttt{GPT-4o} to generate reference responses and evaluated model outputs using the AlpacaEval2 evaluation protocol to compute validation scores. The checkpoint with the highest validation score was selected as the final model.

\paragraph{Evaluation.}
For AlpacaEval2 and ArenaHardV2, we adopted \texttt{GPT-4-1106-preview} as the reference model to generate baseline responses paired with each evaluation prompt. AlpacaEval2 consists of 805 user prompts. Evaluation was conducted in a head-to-head setting, where model outputs were compared against the reference responses and judged by a generative evaluator, resulting in win rates ranging from 0\% to 100\%. We followed the length-controlled win-rate protocol recommended by \cite{dubois2024length} to mitigate length bias. In our implementation, we replaced the default GPT-4 judge with \texttt{GPT-4o}. Notably, we further employed a stronger \texttt{GPT-5} as the final evaluator to ensure robust and reliable assessment. ArenaHardV2 comprises 500 challenging real-world user queries, and we employed the same evaluation protocol as AlpacaEval2. For WildBench, which contains 1,024 user prompts with some multi-turn interactions, we followed the official evaluation protocol and used point-wise scoring instead of head-to-head comparison\footnote{\url{https://github.com/allenai/WildBench/blob/main/evaluation/eval_template.score.v2.md}}. The final scores are reported on a 0-100 scale. For the reasoning benchmarks, we used \texttt{evalscope} to conduct automatic evaluation\footnote{\url{https://github.com/modelscope/evalscope}}. 

\begin{figure}[!t]
    \centering
    \resizebox{\linewidth}{!}{
    \includegraphics[width=\linewidth]{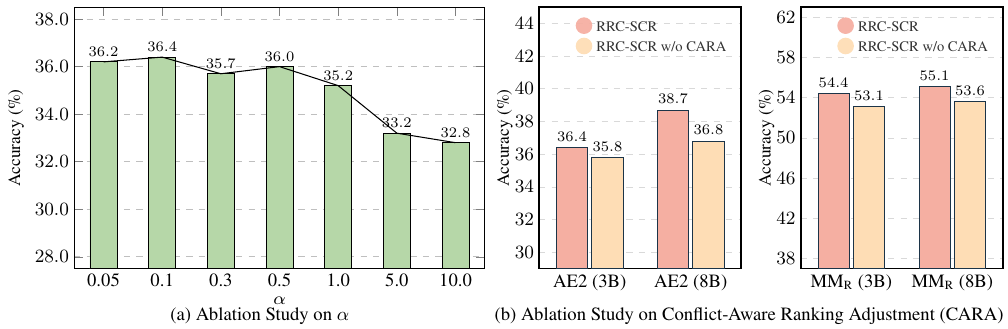}}
    % \vspace{-4mm}
    \caption{
    Ablation Study
    }
    \label{fig:ablation_study}
\end{figure}

\subsection{Ablation Study}
We conduct a comprehensive ablation study to investigate the sensitivity of our method to the reward scaling factor $\alpha$ in SCR. Specifically, as shown in Figure~\ref{fig:ablation_study}(a), we vary $\alpha$ over a wide range from 0.05 to 10.0, and evaluate performance using SCR with 3B-scale reward models. The results indicate that the performance remains relatively stable when $\alpha$ lies in the range of 0.05 to 0.5, suggesting that our method is not highly sensitive to moderate changes in reward scaling. However, when $\alpha$ increases beyond 1.0, we observe a gradual performance degradation. This behavior can be attributed to the normalization mechanism in GRPO-style RL. Since advantages are normalized within each batch, moderate rescaling of reward values does not significantly affect the relative magnitudes of policy gradients. 
In contrast, overly large scaling factors may amplify noise and lead to unstable advantage normalization, which in turn degrades training stability and final performance.

We further investigate the effectiveness of Conflict-Aware Ranking Adjustment (namely CARA) in SCR. As shown in Figure~\ref{fig:ablation_study}(b), removing CARA consistently leads to performance drops across different backbone scales and benchmarks. 
Specifically, on AE2, SCR with CARA outperforms the variant without CARA by 0.6\% and 1.9\% on the 3B and 8B backbones, respectively. 
Similarly, on MM$_\text{R}$, we observe improvements of 1.3\% (3B) and 1.5\% (8B) when CARA is enabled. These results confirm that CARA plays an important role in mitigating the adverse effects of conflicting pairwise preferences produced by generative reward models.

\subsection{Results on Additional Backbone Models}
To evaluate the generality of our approach across different policy backbones, we further conduct experiments on Qwen2.5-7B-Instruct. As shown in Table~\ref{tab:results_on_qwen}, we observe trends highly consistent with those on Qwen2.5-7B-Instruct. In particular, RRC-based methods, including both SCR and AGR variants, consistently outperform DPO, SimPO, and our RL baselines, demonstrating that the effectiveness of RRC is not tied to a specific backbone architecture or model scale.
An interesting observation is that removing explicit \texttt{<think>} traces (\textit{i.e.}, \textit{w/o Thinking}) does not lead to noticeable performance degradation on certain benchmarks, especially on MM$_\text{R}$ and MATH. In some cases, the performance is even slightly improved. We hypothesize that this phenomenon is related to the data distribution used during SFT, where a large portion of reasoning-related samples are embedded within open-ended chat formats rather than strictly structured chain-of-thought annotations.  Similar findings have also been reported in \cite{bhaskar2025language}.

\begin{table}[!t]
    \centering
    \resizebox{0.45\linewidth}{!}{
    \begin{tabular}{lcc}
    \toprule
    \textbf{Dimension} & \textbf{SCR} & \textbf{AGR} \\
    \midrule
    Structure & Fully Connected & Bipartite Graph \\
    Preference Query & $\mathcal{O}(m\cdot \log m)$ & $\mathcal{O}(m \cdot n)$ \\
    Scalability & Limited & Better \\
    \bottomrule
\end{tabular}}
    \vspace{1mm}
    \caption{
    Comparison between SCR and AGR in terms of comparison structure, preference query complexity, and scalability.
    }
    \vspace{-4mm}
    \label{tab:comparison_two_approaches}
\end{table}

\section{Additional Analyses}

\subsection{Behavioral Differences Between SCR and AGR}
\label{app:comparison_our_approaches}
As summarized in Table~\ref{tab:comparison_two_approaches}, SCR and AGR mainly differ in their comparison structures and in the number of preference queries to the generative reward model. SCR performs pairwise comparisons among all sampled responses, which induces a fully connected tournament over the response set. In contrast, AGR compares each response only against a set of anchors, forming a bipartite graph between responses and anchors. This structural difference leads to substantially different computational costs. We further compare the number of queries to the generative reward model required by the two approaches. SCR requires pairwise comparisons between all response pairs, resulting in
\begin{equation}
N_{\text{SCR}} = \tfrac{m(m-1)}{2} = \mathcal{O}(m\cdot\log m)
\end{equation}
while AGR only performs response--anchor comparisons,
\begin{equation}
N_{\text{AGR}} = m \cdot n = \mathcal{O}(mn)
\end{equation}
The difference in query cost is therefore
\begin{equation}
N_{\text{SCR}} - N_{\text{AGR}} = m\!\left(\tfrac{m-1}{2} - n\right)
\end{equation}
which is positive whenever $n < \tfrac{m-1}{2}$, a typical setting in practice.
Up to constant factors, this gap scales on the order of $m(m-n)$ when $n<m$, highlighting the improved scalability of AGR under larger sampling budgets. We also provide the corresponding ranking algorithms in Algorithms~\ref{alg:scr} and~\ref{alg:agr}.

\subsection{Performance vs. Time Cost}

We further analyze the computational overhead introduced by ranking-based rewards and study the trade-off between performance gains and training efficiency.

\paragraph{Efficiency Optimization.}
To reduce the practical overhead of generative reward models, we optimize reward computation from two aspects. First, we deploy generative reward models in parallel using \texttt{vLLM}, where 8 GPUs are used to host 8 model instances for concurrent preference evaluation. Second, we adopt asynchronous reward querying to overlap reward computation with policy optimization. With a sampling size of 8, the additional ranking computation in SCR takes approximately 0.5ms per batch, compared with 0.2ms for discriminative reward models under the same setting. This demonstrates that the overhead introduced by ranking operations remains small at the reward computation level.
For AGR, we further improve efficiency by precomputing anchor responses. Specifically, anchor generation requires approximately 1 hour using 16 GPUs. After anchor construction, AGR training takes around 9 hours, leading to an overall training cost comparable to the baseline RL pipeline.

\begin{table}[t]
\centering
\resizebox{0.7\linewidth}{!}{
\begin{tabular}{lrcc}
\toprule
Method & Training Time & AlpacaEval2 & Gain \\
\midrule
Discriminative RM & 8.2h & 33.2 & - \\
Baseline RL (GRM w/ PRC) & 10.5h & 35.8 & - \\
RRC-SCR & 13.2h & 38.7 & +2.9 \\
RRC-AGR & 10.6h & 39.4 & +3.6 \\
RRC-SCR + voting@8 & 15.2h & 40.0 & +4.2 \\
RRC-AGR + voting@8 & 11.8h & 41.3 & +5.5 \\
\bottomrule
\end{tabular}}
\caption{Comparison of end-to-end training cost and performance on AlpacaEval2 (8B-scale models with Thinking). Improvements are computed over the baseline RL pipeline.}
\label{tab:efficiency}
\end{table}

\begin{table}[!t]
\centering

\begin{tabular}{lcccc}
\toprule
Method & AE2 & WiB & MMR & MATH\\
\midrule
GRM w/ PRC + Majority Voting & 32.8 & 52.1 & 52.8 & 45.2\\
GRM w/ PRC + Soft Aggregation & 30.1 & 51.4 & 48.2 & 44.8\\
GRM w/ RRC-SCR & \textbf{36.4} & 55.6 & 54.4 & 45.6\\
GRM w/ RRC-AGR & 35.8 & \textbf{56.8} & \textbf{55.0} & \textbf{46.4}\\
\bottomrule
\end{tabular}
\caption{Compute-matched comparison between PRC and RRC (3B-scale models with Thinking). All compared methods use the same number of GRM queries as RRC under the same sampling configuration. Note that discriminative reward models are not included because they do not support the same computation scaling mechanism based on repeated generative preference judgments.}
\label{tab:compute_matched}
\end{table}

\paragraph{Performance-Efficiency Analysis.}
Table~\ref{tab:efficiency} summarizes the end-to-end training cost and performance improvements under a fixed hardware budget. Compared with the baseline RL pipeline (GRM w/ PRC), SCR increases the training time from 10.5h to 13.2h, while improving AlpacaEval2 performance from 35.8\% to 38.7\%. When combined with voting@8, SCR further improves performance to 40.0\% with a moderate increase in training cost. In contrast, AGR provides a more favorable performance-efficiency trade-off by leveraging a small set of precomputed anchor responses. With nearly identical training time to the baseline RL pipeline (10.6h vs. 10.5h), AGR improves AlpacaEval2 from 35.8\% to 39.4\%. Moreover, AGR with voting@8 achieves the best performance of 41.3 with only a modest increase in training cost. These results demonstrate that ranking-based reward construction can effectively improve RL training with generative reward models while maintaining practical efficiency.

\paragraph{Compute-Matched Comparison.}
To investigate whether the improvements of RRC mainly come from additional reward model computation, we conduct a compute-matched comparison with a strengthened PRC baseline. Specifically, we replace majority voting in GRM w/ PRC with soft aggregation, where preference probabilities from multiple CoT-enabled GRM queries are averaged to construct continuous rewards. This variant uses a comparable number of GRM evaluations while providing a finer-grained reward signal. As shown in Table~\ref{tab:compute_matched}, soft aggregation does not consistently improve over standard majority voting and remains inferior to RRC across all evaluated benchmarks. For example, on AlpacaEval2, the soft aggregation can achieve 30.1\%, compared with 36.4\% from RRC-SCR and 35.8\% from RRC-AGR. These results suggest that allocating additional inference computation alone is insufficient. The improvement mainly comes from how comparison signals are aggregated: RRC preserves global preference structures and relative margins, whereas score aggregation treats each comparison independently.

\subsection{Point-wise Generative Reward Models}
In this subsection, we further investigate point-wise generative reward models, which directly optimize scalar preference prediction. Specifically, given an input-response pair $(x,y)$, the model is trained to generate a preference label $z$ indicating the quality of the response (\textit{e.g.}, 1, 2, and 3). The training objective is formulated as:
\begin{equation}
\mathcal{L}_{\mathrm{point}} = -\mathbb{E}_{(x,y,z)\sim\mathcal{D}} \log \pi_{\theta}(z|x,y)
\end{equation}
During inference, we use the probability of the positive preference token as the scalar reward:
\begin{equation}
r(x,y)=\pi_{\theta}(z^{+}|x,y)
\end{equation}
where $z^{+}$ denotes the token corresponding to a positive preference.
Although point-wise generative reward models can directly produce scalar preference scores, we find that the resulting reward signals suffer from limited resolution.

To analyze this issue, we train a point-wise generative reward model on the HelpSteer3 dataset and examine the distribution of the final preference label token probabilities under CoT-based generation. As shown in Table~\ref{tab:pointwise_analysis}, the predicted probabilities are highly concentrated near extreme values, with only a small fraction of samples falling into the middle range. This phenomenon indicates scalar compression, where the reward model tends to produce near-deterministic preference predictions and provides limited resolution for distinguishing responses with similar quality. We attribute this behavior to the fact that CoT generation often determines the final preference token, resulting in saturated probability distributions. In contrast, RRC constructs rewards from structured comparisons among responses, explicitly preserving relative ranking and margin information even when individual preference probabilities become saturated. Therefore, the advantage of RRC does not simply come from producing finer-grained scalar rewards, but from better leveraging the comparative nature of generative reward models to provide more informative learning signals for RL. Importantly, the limitation is not caused by the use of token probabilities itself, but by using them as absolute rewards. Even when the underlying preference prediction is accurate, saturated probabilities discard relative differences among responses. RRC avoids this issue by aggregating multiple comparative judgments into structured rankings.

\begin{table}[!t]
\centering
\begin{tabular}{lccc}
\toprule
Probability Range & [0,0.1] & [0.1,0.9] & [0.9,1.0]\\
\midrule
Percentage (\%) & 42.3 & 11.7 & 46.0\\
\bottomrule
\end{tabular}
\caption{Distribution of preference token probabilities from a point-wise generative reward model trained on HelpSteer3.}
\label{tab:pointwise_analysis}
\end{table}

\clearpage

\begin{algorithm}[t]
\caption{Self-Competitive Ranking (SCR) with Majority Voting and CARA}
\label{alg:scr}
\begin{algorithmic}[1]
\REQUIRE Prompt $x$; responses $O=\{o_1,\ldots,o_m\}$; generative reward model $\mathrm{GRM}$; voting budget $V$; scaling factor $\alpha$; flag \texttt{use\_CARA}
\ENSURE Rewards $\{r(x,o_i)\}_{i=1}^m$ and total order $\pi$

\STATE Initialize $w_{ij}\leftarrow 0$ for all $i\neq j$

\FOR{each unordered pair $\{i,j\}$ with $i<j$}
    \STATE $c_{i\succ j}\leftarrow 0$, $c_{j\succ i}\leftarrow 0$
    \FOR{$t=1$ to $V$}
        \STATE Query $\mathrm{GRM}$ on $(x,o_i,o_j)$ and obtain preference
        \IF{$o_i \succ o_j$}
            \STATE $c_{i\succ j}\leftarrow c_{i\succ j}+1$
        \ELSE
            \STATE $c_{j\succ i}\leftarrow c_{j\succ i}+1$
        \ENDIF
    \ENDFOR
    \IF{$c_{i\succ j} \ge c_{j\succ i}$}
        \STATE $w_{ij}\leftarrow c_{i\succ j}$, $w_{ji}\leftarrow c_{j\succ i}$
    \ELSE
        \STATE $w_{ji}\leftarrow c_{j\succ i}$, $w_{ij}\leftarrow c_{i\succ j}$
    \ENDIF
\ENDFOR

\IF{\texttt{use\_CARA} = 1}
    \STATE Initialize partial order $\Omega$ with an empty set
    \WHILE{there exists an undecided pair $(i,j)$}
        \STATE Select $(i,j)$ with the largest $|w_{ij}-w_{ji}|$
        \IF{$w_{ij} \ge w_{ji}$}
            \STATE Add relation $i \prec j$ into $\Omega$
        \ELSE
            \STATE Add relation $j \prec i$ into $\Omega$
        \ENDIF
        \STATE Enforce transitive closure and discard relations forming cycles
    \ENDWHILE
    \STATE Obtain total order $\pi$ by topological sorting of $\Omega$
\ELSE
    \STATE Sort responses by win counts $\sum_{j\neq i}\mathbf{1}[w_{ij} > w_{ji}]$ to obtain $\pi$
\ENDIF

\FOR{each $o_i \in O$}
    \STATE $r(x,o_i) \leftarrow \alpha \cdot |\{o_j \neq o_i : i \prec_\pi j\}|$
\ENDFOR

\RETURN $\{r(x,o_i)\}_{i=1}^m$, $\pi$
\end{algorithmic}
\end{algorithm}

\begin{algorithm}[!t]
\caption{Anchor-Guided Ranking (AGR) with Majority Voting}
\label{alg:agr}
\begin{algorithmic}[1]
\REQUIRE Prompt $x$; responses $O=\{o_1,\ldots,o_m\}$; anchors $A=\{a_1,\ldots,a_n\}$; generative reward model $\mathrm{GRM}$; voting budget $V$; scaling factor $\alpha$
\ENSURE Rewards $\{r(x,o_i)\}_{i=1}^m$

\FOR{each response $o_i \in O$}
    \STATE $s_i \leftarrow 0$
    \FOR{each anchor $a_k \in A$}
        \STATE $c_{i\succ k}\leftarrow 0$, $c_{k\succ i}\leftarrow 0$
        \FOR{$t=1$ to $V$}
            \STATE Query $\mathrm{GRM}$ on $(x,o_i,a_k)$ and obtain preference
            \IF{$o_i \succ a_k$}
                \STATE $c_{i\succ k}\leftarrow c_{i\succ k}+1$
            \ELSE
                \STATE $c_{k\succ i}\leftarrow c_{k\succ i}+1$
            \ENDIF
        \ENDFOR
        \IF{$c_{i\succ k} \ge c_{k\succ i}$}
            \STATE $s_i \leftarrow s_i + 1$
        \ENDIF
    \ENDFOR
    \STATE $r(x,o_i) \leftarrow \alpha \cdot s_i$
\ENDFOR

\STATE Rank responses by descending $r(x,o_i)$
\RETURN $\{r(x,o_i)\}_{i=1}^m$
\end{algorithmic}
\end{algorithm}

\clearpage

\begin{table}[!t]
    \centering
    \resizebox{\linewidth}{!}{

\begin{tabular}{ll}
\toprule[1.1pt]
\textbf{Prompt} & \parbox{16cm}{
Write a rock song, with alternating rhymes [ABAB], about the following ideas below. Avoid common rhymes, jargons, expressions like "neon lights", "shadows creep", "whispers", "air is thick", "bittersweet", "forevermore", "story to be told", etc.\textbackslash n-----\textbackslash n Compassion is the ancient stone bridge, worn smooth by centuries of rain and footsteps, that stretches across the chasm of our own pain. Each stone in its foundation is a memory, a tear, a whispered prayer--each one a testament to the moments when we have stumbled, when we have felt the sharp edges of loss and longing cut into our hearts. And yet, it is this very bridge, this fragile yet enduring connection, that allows us to reach out to others, to understand the weight of their burdens, to offer a hand in the darkness. The bridge does not erase the chasm, but it makes it possible to cross, to find solace in the shared journey, to know that we are not alone in our suffering, and that, together, we can find our way to the other side.} \\  \midrule
\textbf{Baseline (PRC)} & \parbox{16cm}{$<$think$>$\textbackslash nThe user wants a rock song with alternating rhymes (ABAB) about the concept of "compassion" as described in the provided text.\textbackslash n\textbackslash nKey ideas to incorporate:\textbackslash n1.  **Compassion as a bridge:** Connecting people across pain.\textbackslash n2.  **The bridge is ancient, worn:** Smooth, centuries old.\textbackslash n3. ....., showing we care\textbackslash nA fragile, enduring span, that helps us find our way there\textbackslash nThrough the chasm of our suffering, we cross, we cross, we share\textbackslash n\textbackslash n**Verse 2:**\textbackslash n(Guitar: Adds texture, maybe a bridge riff)\textbackslash nEach stone in its foundation tells a story of the fall\textbackslash nOf hopes that shattered like thin ice, of prayers that echoed through the wall\textbackslash nOf moments when the weight was crushing, the edges cut through all\textbackslash nMemories etched into the surface, testifying to how we stood at all\textbackslash n\textbackslash n**Chorus:**\textbackslash n(Reprise with energy)\textbackslash nThis is the weight of the bridge, the weight that we bear\textbackslash nConnecting the sharp edges of our hearts, showing we care\textbackslash nA fragile, enduring span, that helps us find our way there\textbackslash nThrough the chasm of our suffering, we cross, we cross, we share\textbackslash n\textbackslash n**Bridge:**\textbackslash n(Different rhythm or guitar tone.... "connection worn so smooth," "span," "cross."\textbackslash n    *   **Ancient, worn:** "ancient, weathered path," "weathered," "smooth."\textbackslash n    *   **Stones are memories/tears/prayers:** "each stone in its foundation tells a story," "memories etched into the surface," "tears that fell behind," "prayers that echoed."\textbackslash n    *   **Crossing the chasm:** "connecting the sharp edges of our hearts," "find our way there," "helps us find our way," "cross."\textbackslash n    *   **It makes it possible, not erases:** "helps us find our way," "it doesn't mend the cracks," "it provides a hand," "together, we make the journey," "until the dawn is won," "shared weight."\textbackslash n*   **Imagery \& Metaphor:** The song uses strong, evocative images like "weariness," "light is hard to find," "weathered path," "sharp edges," "stones," "tears," "prayers," "chasm," "weight," "span," "journey," "dawn." These help paint a vivid picture of the struggle and the connection.\textbackslash n*   **Rock Song Feel:** The lyrics are written to be sung with energy, especially in the choruses and bridge sections. The language is strong and direct, fitting a rock music style.\textbackslash n$<$/answer$>$"} \\  \midrule
\textbf{Ours (SCR)} & \parbox{16cm}{
$<$think$>$\textbackslash nThis prompt is not simply asking for a song about kindness or emotional support.\textbackslash nIt establishes a **very specific philosophical structure**:\textbackslash n\textbackslash n* Compassion is not a feeling, but an **infrastructure**.\textbackslash n* Pain is not removed, only **made traversable**.\textbackslash n* The bridge is built from **accumulated suffering**, not external rescue.\textbackslash n* Relief comes from **shared movement**, not from victory over pain.\textbackslash n\textbackslash nSo the song must avoid portraying compassion as:\textbackslash n\textbackslash n* a cure,\textbackslash n* a rescue,\textbackslash n* or a triumphant transformation.\textbackslash n\textbackslash nInstead, it should frame compassion as **functional, imperfect, and hard-earned**.\textbackslash n\textbackslash nThat is why the song repeatedly affirms that:\textbackslash n\textbackslash n* the canyon remains,\textbackslash n* the weight remains,\textbackslash n* the damage remains,\textbackslash n\textbackslash nbut movement becomes possible.\textbackslash n\textbackslash nLines like:\textbackslash n\textbackslash n> *"This bridge won't seal the drop below"\textbackslash n> *"That canyon stays, it always will"\textbackslash n\textbackslash ndirectly answer the prompt's insistence that suffering is not erased by connection.\textbackslash nThis preserves the realism of the metaphor instead of turning it into emotional wish-fulfillment.\textbackslash n\textbackslash n---\textbackslash n\textbackslash n\#\# Pain as Construction Material \#\#\# ..... Verse 1\textbackslash n\textbackslash nA canyon cuts the ground in two,\textbackslash nWe hide the breaks we cannot mend,\textbackslash nThe pull of loss keeps dragging through,\textbackslash nLike debts we never get to end.\textbackslash n\textbackslash n\#\#\# Verse 2\textbackslash n\textbackslash nAn aging span of battered stone,\textbackslash nSet down by breath and sleepless nights,\textbackslash nEach block a wound we faced alone,\textbackslash nEach seam still bound by failed goodbyes.\textbackslash n\textbackslash n\#\#\# Verse 3\textbackslash n\textbackslash nA bent-back vow, a scraped-out plea,\textbackslash nA risk we took, a truth we bent,\textbackslash nThey settle in that masonry,\textbackslash nWhere every hurt learns what it meant.\textbackslash n\textbackslash n\#\#\# Verse 4\textbackslash n\textbackslash nThe edges bite, they mark the skin,\textbackslash nThey map the cost of standing fast,\textbackslash nYet every cut we carried in,\textbackslash nBecomes a path across the past.\textbackslash n\textbackslash n\#\#\# Verse 5\textbackslash n\textbackslash nThis bridge won't seal the drop below,\textbackslash nWon't lift the weight we always keep,\textbackslash nBut gives our shaking feet a row,\textbackslash nOf solid ground across the deep.\textbackslash n\textbackslash n\#\#\# Verse 6\textbackslash n\textbackslash nYou cross with me, I cross with you,\textbackslash nNo debts to trade, no flags to raise,\textbackslash nJust hands that know what hands go through,\textbackslash nAnd choose to walk the same hard days.\textbackslash n\textbackslash n\#\#\# Verse 7\textbackslash n\textbackslash nThe load runs light when borne in stride,\textbackslash nThe fall feels less like final ground,\textbackslash nWe trade the need to run and hide,\textbackslash nFor steps that make a steadier sound.\textbackslash n\textbackslash n\#\#\# Verse 8\textbackslash n\textbackslash nPast all the breaks we couldn't close,\textbackslash nPast every word we left unsaid,\textbackslash nWe reach not peace, but still we know...$<$/answer$>$
} \\
\bottomrule[1.1pt]
\end{tabular}}
    \vspace{2mm}
    \caption{
    Case study from the AlpacaEval2 benchmark. Our model produces a more structured and concept-consistent rock song that better aligns with the prompt's philosophical constraints and narrative progression, whereas the baseline tends to rely on surface-level metaphors and generic expressions.
    }
    \label{tab:case_alpacaeval}
\end{table}

\begin{table}[!t]
    \centering
    \resizebox{\linewidth}{!}{

\begin{tabular}{ll}
\toprule[1.1pt]
\textbf{Prompt} & \parbox{16cm}{
create a simple website with a black background that isn't scrollable. There should be a video in the background for the hero and in the center is should say in large bold white text Explore Zeroh. Under the text have a button saying Join Waitlist this button should be faded out and should have a transparent background with a white outline. The text in the button should be white. Once hovering over the button is transitions to a full white button with black text over it.} \\  \midrule
\textbf{Baseline (PRC)} & \parbox{16cm}{$<$think$>$\textbackslash nThe user wants a simple website with a fixed, full-screen background video that doesn't require scrolling. The content should be centered and include:\textbackslash n1.  Large, bold, white text saying "Explore Zeroh".\textbackslash n2.  A "Join Waitlist" button below the text.\textbackslash n    *   The button should be faded out initially.\textbackslash n    *   It should have a transparent background.\textbackslash n    *   It should have a white outline.\textbackslash n    *   The text on the button should be white.\textbackslash n    *   On hover, the button should transition to a full white background.\textbackslash n    *   The text on the button should transition to black.\textbackslash n    *   The hover state should be smooth.\textbackslash n\textbackslash nPlan:\textbackslash n1.  **HTML Structure:** Create a basic HTML5 structure with a $<$!DOCTYPE html$>$ declaration, $<$html$>$, $<$head$>$ for meta tags and styles, and $<$body$>$.\textbackslash n2.  **CSS for Layout \& Overflow:** Use CSS to make the body overflow hidden. Add position: fixed; top: 0; left: 0; width: 100\%; height: 100\%; to the body to make it full-screen.\textbackslash n3.  **CSS for Background Video:** Use a $<$video$>$ element in the $<$body$>$ for the background. Add position: absolute; top: 0; left: 0; width: 100\%; height: 100\%; to the video. Add object-fit: cover; to ensure the video fills the screen and maintains aspect ratio by covering the entire area. Add z-index: -1; so the content sits on top. The video source ($<$source$>$) should be included within the $<$video$>$ tag. Use a sample video URL if none is provided.\textbackslash n4.  **CSS for Content:** Create a container for the exploratory text and the button. Use position: absolute; top: 50\%; left: 50\%; transform: translate(-50\%, -50\%); to center it both horizontally and vertically. Add some basic styling for padding and potentially a background color (initially transparent) if needed for stacking, although z-index and position are the primary stackers here.\textbackslash n5.  **CSS for Header Text:**\textbackslash n    *   Select the header text (.header-text).\textbackslash n    *   Set font-size to a large value (e.g., 50px). You should see the black background, the centered video, the large "Explore Zeroh" text, and the faded-out "Join Waitlist" button. Hovering over the button should make it fully white with black text.\textbackslash n\textbackslash nThis setup provides a simple, fixed, non-scrollable layout with the requested visual effects and centering.$<$/answer$>$"} \\  \midrule
\textbf{Ours (AGR)} & \parbox{16cm}{
$<$think$>$1. Make the page non-scrollable (`overflow: hidden`) and full-screen.\textbackslash n2. Put a full-bleed **background video** behind everything (`position: fixed; object-fit: cover`).\textbackslash.....\textbackslash n\textbackslash n```html\textbackslash n$<$!doctype html$>$\textbackslash n$<$html lang="en"$>$\textbackslash n  $<$head$>$\textbackslash n    $<$meta charset="UTF-8" /$>$\textbackslash n    $<$meta name="viewport" content="width=device-width, initial-scale=1.0" /$>$\textbackslash n    $<$title$>$Explore Zeroh$<$/title$>$\textbackslash n    $<$style$>$\textbackslash n      :root {\textbackslash n        --bg: \#000;\textbackslash n        --white: \#fff;\textbackslash n        --black: \#000;\textbackslash n      }\textbackslash n\textbackslash n      /* Prevent scrolling + full-screen layout */\textbackslash n      html, body {\textbackslash n        width: 100\%;\textbackslash n        height: 100\%;\textbackslash n        margin: 0;\textbackslash n        overflow: hidden; /* not scrollable */\textbackslash n        background: var(--bg);\textbackslash n        font-family: ui-sans-serif, system-ui, -apple-system, Segoe UI, Roboto, Helvetica, Arial;\textbackslash n      }\textbackslash n\textbackslash n      /* Background video */\textbackslash n      .bg-video {\textbackslash n        position: fixed;\textbackslash n        inset: 0;\textbackslash n        width: 100\%;\textbackslash n        height: 100\%;\textbackslash n        object-fit: cover;\textbackslash n        z-index: 0;\textbackslash n        background: var(--bg);\textbackslash n      }\textbackslash n\textbackslash n      /* Optional dark overlay to keep text readable */\textbackslash n      .overlay {\textbackslash n        position: fixed;\textbackslash n        inset: 0;\textbackslash n        z-index: 1;\textbackslash n        background: rgba(0, 0, 0, 0.45);\textbackslash n      }\textbackslash n\textbackslash n      /* Centered hero content */\textbackslash n      .hero {\textbackslash n        position: relative;\textbackslash n        z-index: 2;\textbackslash n        width: 100\%;\textbackslash n        height: 100\%;\textbackslash n        display: flex;\textbackslash n        align-items: center;\textbackslash n        justify-content: center;\textbackslash n        text-align: center;\textbackslash n        padding: 24px;\textbackslash n        box-sizing: border-box;\textbackslash n      }\textbackslash n\textbackslash n      .hero-inner {\textbackslash n        display: flex;\textbackslash n        flex-direction: column;\textbackslash n        gap: 18px;\textbackslash n        align-items: center;\textbackslash n      }\textbackslash n\textbackslash n      h1 {\textbackslash n        margin: 0;\textbackslash n        color: var(--white);\textbackslash n        font-weight: 800;\textbackslash n        letter-spacing: 0.5px;\textbackslash n        line-height: 1.05;\textbackslash n        font-size: clamp(42px, 6vw, 86px);\textbackslash n        text-shadow: 0 10px 30px rgba(0,0,0,0.35);\textbackslash n      }\textbackslash n\textbackslash n      /* Faded-out outlined button */\textbackslash n      .btn {\textbackslash n        appearance: none;\textbackslash n        border: 2px solid rgba(255, 255, 255, 0.9);\textbackslash n        background: transparent;\textbackslash n        color: var(--white);\textbackslash n        font-weight: 700;\textbackslash n        font-size: 16px;\textbackslash n        padding: 12px 22px;\textbackslash n        border-radius: 999px;\textbackslash n        cursor: pointer;\textbackslash n\textbackslash n        /* "faded out" look */\textbackslash n        opacity: 0.65;\textbackslash n\textbackslash n        /* smooth hover transition */\textbackslash n        transition: background-color 220ms ease, color 220ms ease, opacity 220ms ease, transform 220ms ease;\textbackslash n        will-change: background-color, color, opacity, transform;\textbackslash n      }\textbackslash n\textbackslash n      .btn:hover {\textbackslash n        background: var(--white);\textbackslash n        color: var(--black);\textbackslash n        opacity: 1;\textbackslash n        transform: translateY(-1px);\textbackslash n      }\textbackslash n\textbackslash n      .btn:active {\textbackslash n        transform: translateY(0px);\textbackslash n      }\textbackslash n\textbackslash n      .btn:focus-visible {\textbackslash n        outline: 3px solid rgba(255,255,255,0.45);\textbackslash n        outline-offset: 4px;\textbackslash n        opacity: 1;\textbackslash n      }\textbackslash n\textbackslash n      /* Respect reduced motion preferences */\textbackslash n      @media (prefers-reduced-motion: reduce) {\textbackslash n        .btn { transition: none; }\textbackslash n      }\textbackslash n    $<$/style$>$\textbackslash n  $<$/head$>$\textbackslash n\textbackslash n  $<$body$>$\textbackslash n    $<$!-- Replace src with your own video file URL/path --$>$\textbackslash n    $<$video class="bg-video" autoplay muted loop playsinline$>$\textbackslash n  ....$<$/answer$>$
} \\
\bottomrule[1.1pt]
\end{tabular}}
    \vspace{2mm}
    \caption{
    Case study from the ArenaHardV2 benchmark. We use \texttt{GPT-5} as an automatic judge to evaluate the correctness of the generated code. Compared with the baseline, the AGR-trained model produces code with fewer functional and stylistic errors, and more faithfully satisfies the prompt's layout and interaction constraints.
    }
    \label{tab:case_arenahard}
\end{table}

\end{document}